\documentclass[letterpaper]{article} 
\usepackage{aaai2026}  
\usepackage{times}  
\usepackage{helvet}  
\usepackage{courier}  
\usepackage[hyphens]{url}  
\usepackage{graphicx} 
\usepackage{natbib}  
\usepackage{caption} 
\usepackage{algorithm}
\usepackage{algorithmic}
\usepackage{multirow}
\usepackage{booktabs}
\usepackage{amsmath}
\usepackage{url}
\usepackage{xcolor}
\usepackage{colortbl}
\usepackage{threeparttable}
\usepackage{tcolorbox}
\usepackage{tabularx}
\usepackage{amssymb}
\usepackage{subcaption} 

\usepackage{newfloat}
\usepackage{listings}
\DeclareCaptionStyle{ruled}{labelfont=normalfont,labelsep=colon,strut=off} 
\floatstyle{ruled}
\newfloat{listing}{tb}{lst}{}
\floatname{listing}{Listing}
\title{Judge Q: Trainable Queries for Optimized Information Retention\\ in KV Cache Eviction}
\author{
    Yijun Liu,
    Yixuan Wang,
    Yuzhuang Xu,
    Shiyu Ji, \\
    Yang Xu,
    Qingfu Zhu,
    Wanxiang Che\thanks{Corresponding author}
}
\affiliations{
    \textsuperscript{\rm}Research Center for Social Computing and Interactive Robotics, \\ Harbin Institute of Technology, Harbin, China\\
    \{yijunliu, car\}@ir.hit.edu.cn
}

\usepackage{bibentry}

\begin{document}

\maketitle

\begin{abstract}
Large language models (LLMs) utilize key-value (KV) cache to store historical information during sequence processing.
The size of KV cache grows linearly as the length of the sequence extends, which seriously affects memory usage and decoding efficiency. 
Current methods for KV cache eviction typically utilize the last window from the pre-filling phase as queries to compute the KV importance scores for eviction. 
Although this scheme is simple to implement, it tends to overly focus on local information, potentially leading to the neglect or omission of crucial global information.
To mitigate this issue, we propose \textbf{Judge Q}, a novel training method which incorporates a soft token list.
This method only tunes the model’s embedding layer at a low training cost. 
By concatenating the soft token list at the end of the input sequence, we train these tokens' attention map to the original input sequence to align with that of the actual decoded tokens.
In this way, the queries corresponding to the soft tokens can effectively capture global information and better evaluate the importance of the keys and values within the KV cache, thus maintaining decoding quality when KV cache is evicted.
Under the same eviction budget, our method exhibits less performance degradation compared to existing eviction approaches.
We validate our approach through experiments conducted on models such as Llama-3.1-8B-Instruct and Mistral-7B-Instruct-v0.3, using benchmarks including LongBench, RULER, and Needle-in-a-Haystack.
Results indicate an improvement of approximately 1 point on the LongBench and over 3 points on RULER.
This proposed methodology can be seamlessly integrated into existing open-source models with minimal training overhead, thereby enhancing performance in KV cache eviction scenarios.
\end{abstract}

\begin{links}
    \link{Code}{https://github.com/Mambaaaaaaaaa/Judge-Q}
\end{links}

\section{Introduction}

Large language models (LLMs) have demonstrated remarkable performance across a wide range of domains, including language modeling~\cite{hendrycks2021measuringmassivemultitasklanguage, suzgun2022challengingbigbenchtaskschainofthought}, text comprehension~\cite{rajpurkar-etal-2016-squad, 2017arXivtriviaqa}, and code understanding~\cite{chen2021evaluatinglargelanguagemodels, jain2024livecodebenchholisticcontaminationfree}. 
Currently, many scenarios~\cite{liu2025comprehensive, liu2025surveytransformercontextextension, chen2025towards, lu2025uni, lu2025out} involve input sequences with over ten thousand tokens, such as books, code repositories~\cite{kočiský2017narrativeqareadingcomprehensionchallenge, zhang2023repocoderrepositorylevelcodecompletion}, and dialogue systems with long-term histories~\cite{li2024streamingdialogueprolongeddialoguelearning}.
LLMs employ KV cache to store historical information during the pre-filling stage, and reuse them during decoding. 
The size of KV cache increases linearly with input sequence length. 
In long sequence scenarios, KV cache often consumes substantial memory space, posing a significant bottleneck for inference, particularly in resource-limited environments.

\begin{figure}
    \centering
    \includegraphics[width=\linewidth]{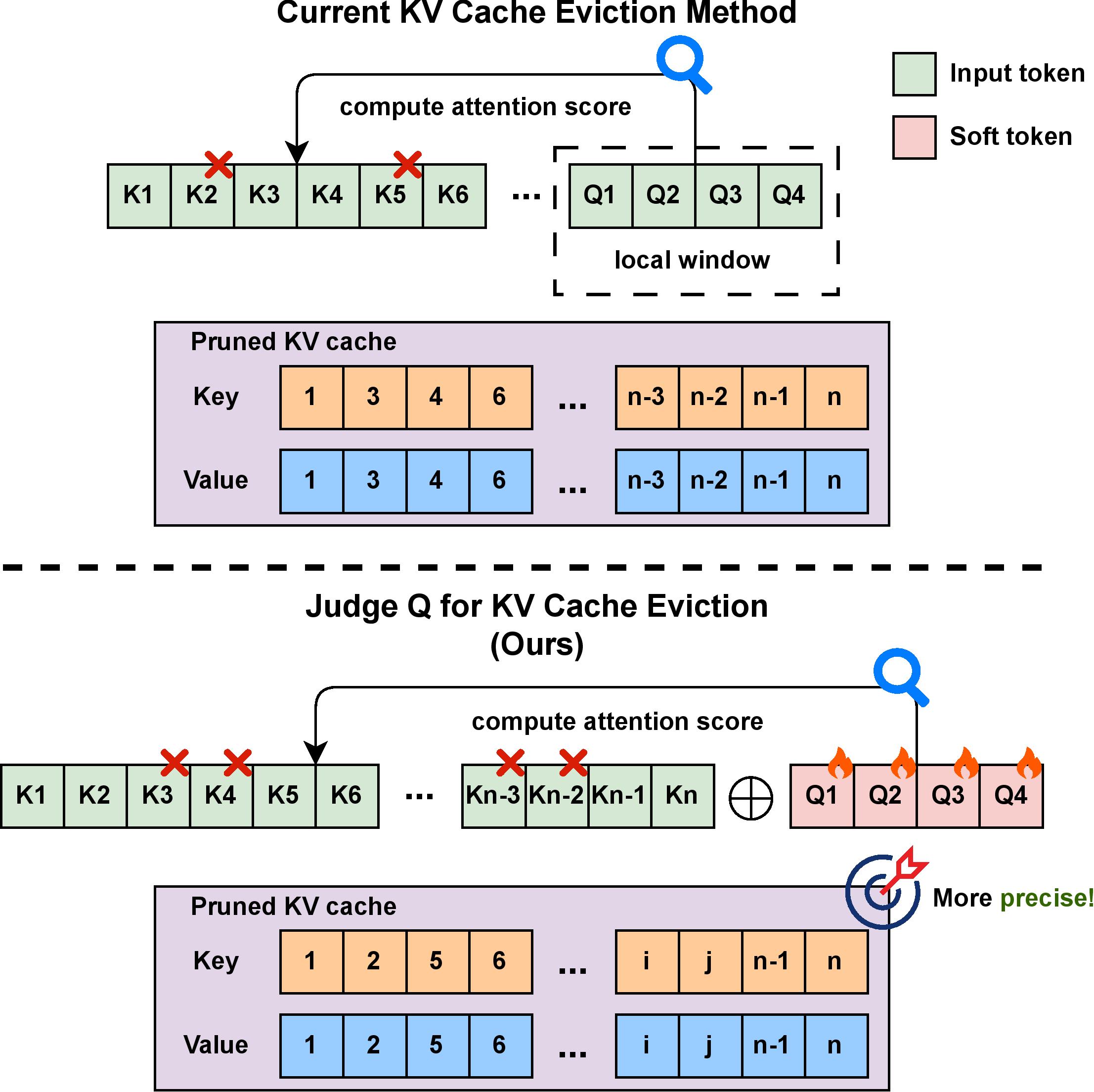}
    \caption{Difference between current KV cache eviction methods and ours when  calculating the importance score of key-value pairs during prefilling stage. We utilize trainable soft tokens to effectively leverage global information.}
    \label{fig:intro}
\end{figure}

To tackle the challenge of excessive memory consumption by KV cache, a series of KV cache eviction methods~\citep{zhang2023h2oheavyhitteroracleefficient, li2024snapkvllmknowslooking, cai2024pyramidkv} are proposed from the perspective of directly reducing the elements within KV cache during the pre-filling stage. 
~\citet{li2024snapkvllmknowslooking} observe that only a small portion of tokens in the prompt have high attention scores which contain crucial information to generate satisfactory responses. 
These strategies leverage the characteristics of auto-regressive language models and the observation that questions often appear towards the end of inputs. 
They typically choose queries from the last window during pre-filling to calculate the importance scores of key-value pairs based on attention.
By retaining the top-k most important portions and discarding the rest, they effectively manage the size of the KV cache.

This straightforward approach of directly selecting the last window is simple to implement, however, it has limitations.
Firstly, it tends to overly focus on local information within the last window while neglecting global content, which can result in potential information loss.
Additionally, if the last window does not contain question, the performance of these methods often significantly degrades.
In fact, the primary objective of KV cache eviction is to reduce memory usage while maintaining  decoding quality. 
Therefore, identifying the key-value pairs that are crucial for decoding the subsequent tokens is logical and ideal.

In this paper, we discover that directly using the original decoded tokens to select critical key-value pairs preserves high decoding quality, representing the theoretical upper bound of KV cache eviction methods.
However, during the prefilling stage, the actual decoded tokens are not available. 
Inspired by this phenomenon, we propose a training method called \textbf{Judge Q}. 
This approach involves introducing a list of soft tokens into the model's vocabulary and leveraging them to perform KV cache eviction during the pre-filling stage. 
Specifically, during the training phase, we concatenate these soft tokens directly to the end of original input sequence. 
The soft tokens are trained to align the attention map to input sequence with that of the actual decoded tokens, enabling the soft tokens to capture global information effectively. 
This process incurs minimal training costs, as it involves training only the parameters associated with the soft tokens in the model's embedding layer, while keeping the rest of the model's weights frozen.
During inference, we similarly append these soft tokens to the end of the original input sequence in the pre-filling stage.
The queries corresponding to soft tokens are used to calculate the importance scores of the key-value pairs in KV cache across the entire input sequence. 
Based on these scores, we retain the top-k key-value pairs with the highest scores and discard the rest.
After the pre-filling stage, the soft tokens are also removed. 
The pruned KV cache is then used for the subsequent decoding process.
We conduct experiments using two models, Llama-3.1-8B-Instruct~\citep{dubey2024llama} and Mistral-7B-Instruct-v0.3~\cite{jiang2023mistral7b}, trained with self-generated data from the ShareGPT dataset~\cite{ShareGPT-Chinese-English-90k}. 
Our method is evaluated on LongBench~\cite{bai2023longbench}, RULER~\cite{hsieh2024ruler}, and Needle-in-a-Haystack~\cite{kamradt2023needle}. 
Compared to other methods performing KV cache eviction during the pre-filling stage, our approach consistently achieves superior results under the same budget and more closely approaches the theoretical upper bound of using actual decoded tokens.

The main contributions of this paper are as follows:
\begin{itemize}
    \item We propose Judge Q, a novel training method that alleviates the limitations of existing KV cache eviction methods, which tend to focus excessively on local information while overlooking  global content during pre-filling.
    \item The proposed method can be adapted to any open-source model with minimal training costs. By fine-tuning only the embedding layer of the model, it enhances the model's performance in KV cache eviction scenarios.
    \item Experimental results on three benchmarks demonstrate that our method outperforms existing KV cache eviction methods during the pre-filling stage under the same budget, with significant  gains in low-resource settings. Notably, our approach achieves nearly 1 point improvement on LongBench and over 3 points on RULER.
\end{itemize}

\section{Related Works}

\subsection{KV Cache Eviction}
To address the challenge of reducing the memory footprint of KV cache, a series of methods are proposed from the perspective of reducing the number of elements in KV cache.
These strategies can effectively control KV cache size by evaluating the importance of key-value pairs within the input sequence during pre-filling, retaining only those with high importance scores and discarding less critical ones.
This approach is grounded in the observation of the sparsity of attention~\cite{zhang2023h2oheavyhitteroracleefficient} and that only a small portion of tokens convey essential information for response generation during the pre-filling stage~\cite{li2024snapkvllmknowslooking}.
Most of these methods compute the cumulative attention scores of key-value pairs with respect to queries within the local window during the prefilling stage, which are used as the corresponding importance scores for eviction.
Methods like H2O~\cite{zhang2023h2oheavyhitteroracleefficient} and Scissorhands~\cite{liu2023scissorhands} perform dynamic KV cache eviction, while approaches ~\cite{li2024snapkvllmknowslooking, chen2024nacl, he2025treekv} conduct a static eviction.
Additionally, some work~\cite{feng2024ada, cai2024pyramidkv, yang2024pyramidinfer} focuses on dynamically allocating budgets to optimize memory usage.
Although these methods achieve effective KV cache eviction during pre-filling stage.
As shown in Fig~\ref{fig:intro},their directly relying on the queries of local window to score the importance of key-value pairs often results in an overemphasis on local information while neglecting global content, thereby limiting performance.
Therefore, it is essential to develop a KV cache eviction method that can retain more information during the prefilling stage to support subsequent generation effectively.

Besides the mainstream KV cache eviction methods mentioned above, there are also other KV cache eviction approaches from different perspectives.
~\citet{devoto2024simple} use the L2 norm of the key vector as the importance score of key-value pair.
~\citet{ge2023model} adaptively select and apply different eviction strategies for different layers and heads.
Lookahead Q-Cache~\cite{wang2025lookahead} utilizes decoded pseudo queries to guide KV cache eviction.
And~\citet{xiao2023efficient} propose StreamingLLM, which only retains the tokens from the beginning and the local segments.
While~\citet{xiao2024duoattention} train the heads of model to selectively utilize streaming attention and full attention.

\subsection{KV Cache Compression}
In addition to KV cache eviction methods, memory space can also be conserved by compressing the representation of KV cache.
~\citet{xu2024think} identify substantial redundancy in the channel dimension of KV cache, and propose THINK, which reduces memory usage by retaining only the essential channels of KV cache vectors.
Besides direct channel discarding for dimension reduction, some methods~\citep{singhania2024loki, zhang2024lorc, chang2025palu} utilize techniques such as Singular Value Decomposition (SVD) and Principal Component Analysis (PCA) to effectively reduce the dimension of key and value vectors in KV cache.

Moreover,~\citet{bolya2022token, wan2024d2o}focus on merging KV cache representations at token level, enabling key-value vectors to store more information per element, thus shrinking KV cache size.
Furthermore, several studies~\cite{sun2024you, yang2024kvsharer, wu2024layer} exploit the similarities of KV cache representations across different layers and employ cross-layer KV cache sharing to reduce its memory footprint.

\begin{figure}
    \centering
    \includegraphics[width=\linewidth]{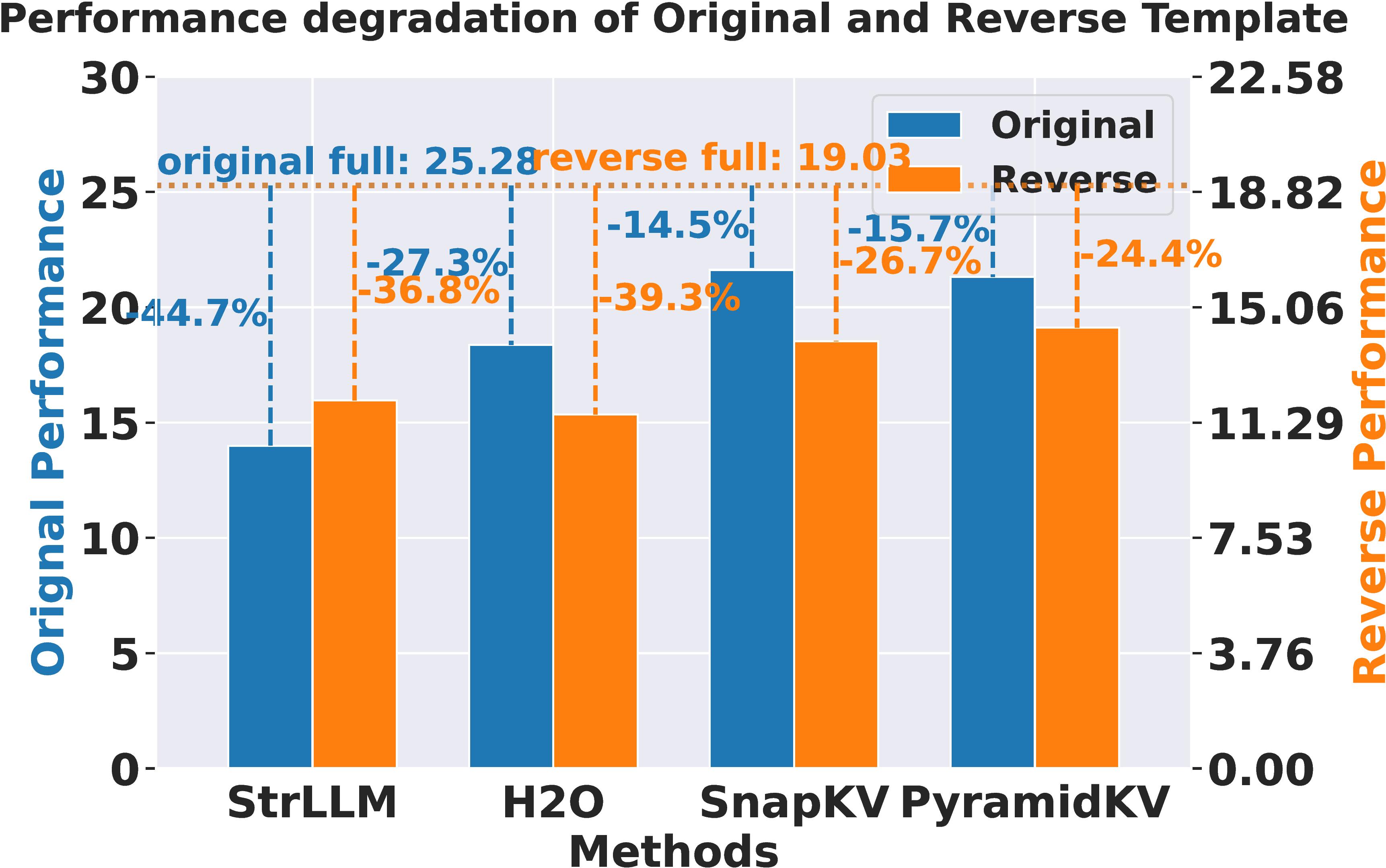}
    \caption{Performance degradation of different pruning methods on Single-QA tasks before and after prompt adjustment, with all methods having a KV budget of 512 tokens.}
    \label{fig:reverse_qa}
\end{figure}

\section{Observations}
\label{sec:obs}
Although existing KV cache eviction methods achieve memory reduction at the expense of small performance degradation, our empirical analysis reveals several fundamental limitations that constrain their practical effectiveness.

\paragraph{(1) Directly utilizing queries from the last window for importance calculation may lead to information loss.}
Current KV cache eviction methods during the pre-filling phase are predominantly designed based on the observation that the problem contexts tend to occur at the end of inputs.
These methods typically compute importance scores for KV cache using the queries corresponding to the tokens in the last window. 
However, they are problem-specific and do not effectively preserve essential information. 
When the last window does not contain the question, such methods often experience significant performance degradation.
To evaluate this limitation, we conduct experiments using Single-QA tasks from LongBench, systematically modifying the prompt template by moving the question components from their conventional terminal position to the initial segment. 
Meanwhile, the prompt is adjusted to ensure semantic coherence. 
Under this modified setting, we evaluate the relative performance degradation of existing eviction methods before and after the prompt adjustment.
As illustrated in Figure~\ref{fig:reverse_qa}, the performance degradation of these pruning methods becomes more serious after prompt adjustment. 
This indicates that these methods suffer from information loss during importance calculation and are unable to adequately capture and process global contextual information.
\paragraph{(2) KV cache eviction guided by queries corresponding to actual decoded tokens yields excellent results.}
On the other hand, our findings suggest that directly using the queries corresponding to the actual decoded tokens to compute the importance scores of key-value pairs in KV cache for pruning can achieve highly effective results.
We evaluate on the LongBench with Llama-3.1-8B-Instruct, leveraging FullKV-generated responses to guide KV cache eviction.
The results are compared against other eviction methods.
The experimental results, as shown in Table~\ref{tab:response_driven}, demonstrate that under the same budget, the pruning method based on the actual response outperforms other pruning methods. 
This indicates that the decoded tokens are better at capturing global information.
This observation is also logically consistent: since the pruned KV cache is ultimately used for decoding subsequent tokens, directly retaining the key-value pairs most important to the decoded tokens yields highly effective results, representing the theoretical upper bound of KV cache pruning methods during the pre-filling stage.
However, during the pre-filling stage, the content of decoded tokens is not yet known in advance. 
Therefore, it is essential to develop an approximate pruning method that can utilize tokens from the pre-filling stage to identify critical key-value pairs that closely align with those selected based on actual decoded tokens.

\begin{table}
\centering
  \begin{tabularx}{\linewidth}{l *{3}{>{\centering\arraybackslash}X}}
  \toprule
  \multirow{2}{*}[-0.5ex]{\textbf{Method}} & \multicolumn{3}{c}{\textbf{KV Cache Budget}} \\
  \cmidrule(){2-4}
     & 128 & 256 & 512 \\
  \midrule
      Full KV-gen & \textbf{38.27} & \textbf{39.29} & \textbf{39.80} \\
  StreamingLLM & 30.50 & 31.79 & 32.64 \\
  H2O & 33.67 & 34.27 & 35.30 \\
  SnapKV & 34.31 & 36.56 & 38.31 \\
  PyramidKV & 34.08 & 36.00 & 37.58 \\
  \bottomrule
  \end{tabularx}
    \caption{The performance of different pruning methods on LongBench using Llama-3.1-8B-Instruct, all methods are under the same settings in Sec~\ref{sec:experiments}. The result of Full KV is 41.23. The optimal results are highlighted in bold.}
  \label{tab:response_driven}
\end{table}

\section{Method}
\begin{figure*}
    \centering
    \includegraphics[width=\linewidth]{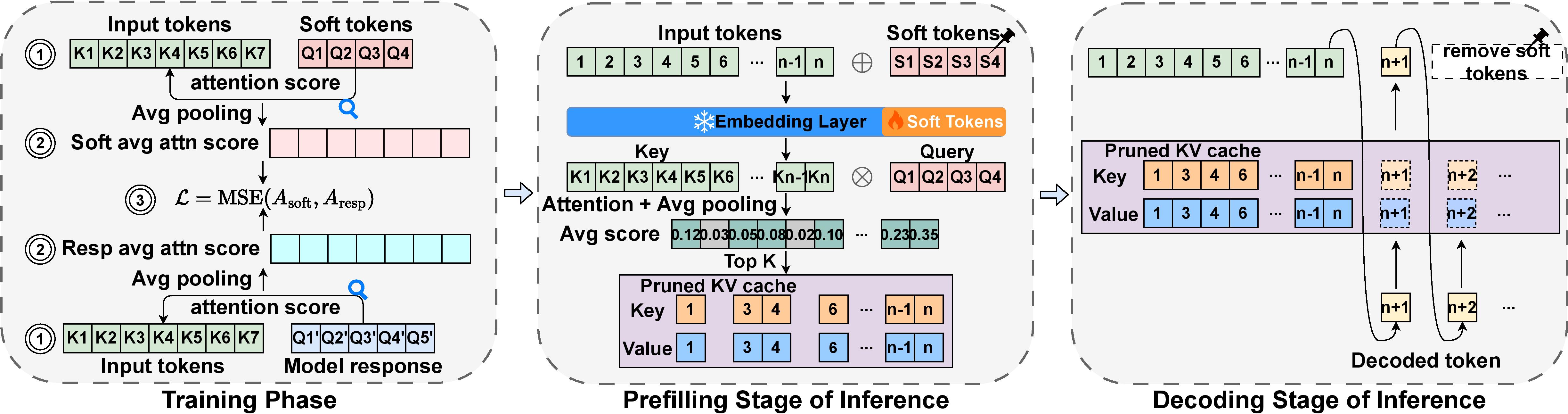}
    \caption{The framework of Judge Q. We specifically show the details of training, prefilling, and decoding stages. The number of soft tokens in the figure is set {n} = 4.}
    \label{fig:framework}
\end{figure*}

\subsection{Framework of Judge Q}
Motivated by the shortcomings of existing KV cache eviction methods during the pre-filling stage mentioned in Section~\ref{sec:obs}, we propose the framework ``Judge Q'' to address these issues.
The framework introduces a set of learnable soft tokens into model's vocabulary, requiring fine-tuning of the embedding layer.
It is illustrated in Figure~\ref{fig:framework}. 
The training procedure is carefully designed to equip these tokens with enhanced global information awareness, while corresponding modifications during inference ensure these tokens actively guide optimized KV cache eviction.

\subsection{Training Model With Soft Tokens}
\label{subsec:training}

Before training, we define the number of soft tokens as $n$ and expand the model’s vocabulary by appending these tokens. 
During the training process, we exclusively fine-tune the subset of parameters in the model's embedding layer that correspond to these soft tokens, while keeping the remaining parameters frozen. 
We construct our training set using 50,000 samples from the ShareGPT dataset, with processing details provided in the repository.
Each sample contains two fields: prompt and response, where the response represents the model's actual output for the given prompt. 

To train the soft tokens to effectively capture global information, we optimize their attention patterns to align with those of the response tokens.
Specifically, the soft tokens and the response tokens are individually concatenated with the prompt tokens:

\begin{equation}
\label{eq:input_soft}
\mathbf{Input}_{\text{soft}}
=
\operatorname{Concat}\big(\mathbf{Prompt},\, \mathbf{Soft}\big)
\end{equation}

\begin{equation}
\label{eq:input_resp}
\mathbf{Input}_{\text{resp}}
=
\operatorname{Concat}\big(\mathbf{Prompt},\, \mathbf{Resp}\big)
\end{equation}
Next, we calculate their attention scores with respect to the prompt, and average these attention scores across the token dimension to derive the attention map.
The training objective is to minimize the Mean Squared Error (MSE) between these two attention maps.
\begin{equation}
\label{eq:att_map_soft}
\mathbf{A}_{\text{soft}} 
= \frac{1}{n} \sum_{i=1}^{n} 
\operatorname{Attention}(\mathbf{Prompt}, \mathbf{Soft}_i)
\end{equation}
\begin{equation}
\label{eq:att_map_gold}
\mathbf{A}_{\text{resp}} 
= \frac{1}{m} \sum_{j=1}^{m} 
\operatorname{Attention}(\mathbf{Prompt}, \mathbf{Resp}_j)
\end{equation}
\begin{equation}
\label{eq:att_loss}
\mathcal{L}
=
\operatorname{MSE}\big(\mathbf{A}_{\text{soft}}, \mathbf{A}_{\text{resp}}\big)
=
\frac{1}{d}\big\lVert\, \mathbf{A}_{\text{soft}} - \mathbf{A}_{\text{resp}} \,\big\rVert_2^2
\end{equation}

This approach ensures that the soft tokens are trained to focus on prompt regions critical for generation, achieving more informed KV cache eviction during inference.

\subsection{Optimizing KV Cache Eviction via Soft Tokens during Inference}

During inference, we leverage the trained soft tokens to perform KV cache eviction. 
Similarly to training procedure, we append $n$ soft tokens sequentially to the end of original input. 
In the pre-filling stage, we compute the attention map of soft tokens with respect to the input, as formulated in Eq~\ref{eq:att_map_soft}. 
Based on this attention map, we retain the key-value pairs corresponding to the top-k highest scores while discarding the rest.
The soft tokens act as probes, which are removed after this pruning step, and the model continues decoding from the end of the original input using the pruned KV cache.

\section{Experiments}
\label{sec:experiments}
\begin{table*}[!t]
\small

\setlength{\tabcolsep}{1mm}
\centering
\begin{threeparttable}

\begin{tabular}{llccccccccccccccccc}
\specialrule{1pt}{0pt}{2pt}
&\multirow{4}{*}{\textbf{~~~Method}} & \multicolumn{3}{c}{\textbf{Single-Doc QA}} & \multicolumn{3}{c}{\textbf{Multi-Doc QA}}& \multicolumn{3}{c}{\textbf{Summ.}}& \multicolumn{3}{c}{\textbf{Few-shot}}& \multicolumn{2}{c}{\textbf{Synth.}} & \multicolumn{2}{c}{\textbf{Code}}&\multirow{4}{*}{\textbf{~~~Avg.}} \\
\cmidrule(lr){3-5}\cmidrule(lr){6-8}\cmidrule(lr){9-11}\cmidrule(lr){12-14}\cmidrule(lr){15-16}\cmidrule(lr){17-18}
&& \rotatebox[origin=c]{45}{\textbf{NQA}} & \rotatebox[origin=c]{45}{\textbf{Qasp}} & \rotatebox[origin=c]{45}{\textbf{MF}} & \rotatebox[origin=c]{45}{\textbf{HQA}} & \rotatebox[origin=c]{45}{\textbf{2Wiki}} & \rotatebox[origin=c]{45}{\textbf{Musq}} & \rotatebox[origin=c]{45}{\textbf{GovR}} & \rotatebox[origin=c]{45}{\textbf{QM}} & \rotatebox[origin=c]{45}{\textbf{MN}} & \rotatebox[origin=c]{45}{\textbf{TREC}} & \rotatebox[origin=c]{45}{\textbf{Triv}} & \rotatebox[origin=c]{45}{\textbf{SAM}} & \rotatebox[origin=c]{45}{\textbf{PCnt}} & \rotatebox[origin=c]{45}{\textbf{PRe}} & \rotatebox[origin=c]{45}{\textbf{Lcc}} & \rotatebox[origin=c]{45}{\textbf{RB-P}} \\

\specialrule{1pt}{2pt}{2pt}

\multirow{19}{*}{\rotatebox[origin=c]{90}{\textbf{Llama-3.1-8B-Instruct}}}

&~~~Full KV & 31.85 & 15.59 & 28.39 & 29.93 & 22.98 & 18.20 & 34.18 & 23.79 & 27.14 & 72.50 & 92.14 & 43.66 & 8.45 & 97.59 & 61.36 & 51.88 & 41.23 \\
\cline{2-19}

& \multicolumn{18}{c}{\cellcolor{lightgray!25} 
  \textbf{KV Cache Size = 128}
} \\

&~~~StrLLM & 15.35 & 4.96 & 14.83 & 14.68 & 17.76 & 7.55 & 20.56 & 19.66 & 20.55 & 42.50 & 83.12 & 32.14 & \textbf{10.34} & \textbf{96.58} & 46.78 & 40.56 & 30.50 \\

&~~~H2O & 25.11 & 7.00 & 18.28 & 18.15 & \textbf{20.05} & 9.21 & 22.06 & 22.71 & 21.49 & 40.00 & 91.05 & 40.50 & 7.76 & 93.54 & 55.44 & 46.39 & 33.67 \\

&~~~SnapKV & 24.14 & 7.59 & 21.56 & 17.74 & 19.25 & 10.15 & 19.91 & 22.13 & 20.40 & 48.50 & 91.63 & 40.26 & 10.00 & 93.70 & 55.29 & \textbf{46.75} & \underline{34.31} \\

&~~~PyraKV & 25.39 & 6.89 & 21.59 & 16.67 & 17.66 & 8.85 & 20.29 & 22.45 & 20.37 & 50.50 & 89.92 & 40.41 & 9.25 & 93.46 & 55.43 & 46.20 & 34.08 \\

&~~~\textbf{Ours} & \textbf{28.21} & \textbf{10.45} & \textbf{24.47} & \textbf{19.51} & 18.38 & \textbf{12.13} & \textbf{23.85} & \textbf{23.94} & \textbf{22.97} & \textbf{51.50} & \textbf{92.20} & \textbf{41.58} & 9.52 & 92.44 & \textbf{57.00} & 46.22 & \textbf{35.90 }\\

\cline{2-19}
& \multicolumn{18}{c}{\cellcolor{lightgray!25} 
  \textbf{KV Cache Size = 256}
} \\

&~~~StrLLM & 16.57 & 5.61 & 15.74 & 15.34 & 16.10 & 6.96 & 22.96 & 19.69 & 22.99 & 55.50 & 83.18 & 32.54 & \textbf{10.00} & 95.55 & 48.18 & 41.77 & 31.79 \\

&~~~H2O & 25.89 & 8.64 & 19.57 & 17.34 & 18.19 & 10.02 & 23.35 & 22.95 & 23.00 & 41.50 & 91.29 & 41.72 & 7.96 & 94.57 & 56.87 & 46.98 & 34.37 \\

&~~~SnapKV & 27.77 & 10.12 & 22.94 & \textbf{22.48} & 19.84 & 11.15 & 22.64 & 22.94 & 23.23 & 58.00 & 91.32 & 41.46 & 8.20 & 95.42 & \textbf{59.22} & \textbf{48.30} & \underline{36.56} \\

&~~~PyraKV & 26.52 & 8.55 & 22.66 & 17.94 & 18.56 & 9.95 & 22.34 & 23.16 & 22.52 & \textbf{59.00} & 91.69 & 41.35 & 9.35 & \textbf{96.93} & 58.33 & 47.18 & 36.00 \\

&~~~\textbf{Ours} & \textbf{29.76} & \textbf{12.87} & \textbf{25.59} & 22.02 & \textbf{21.76} & \textbf{13.82} & \textbf{26.07} & \textbf{23.53} & \textbf{24.46} & \textbf{59.00} & \textbf{92.15} & \textbf{41.84} & 7.76 & 95.94 & 58.56 & 47.84 & \textbf{37.69} \\

\cline{2-19}
& \multicolumn{18}{c}{\cellcolor{lightgray!25} 
  \textbf{KV Cache Size = 512}
} \\

&~~~StrLLM & 17.99 & 6.38 & 17.60 & 13.97 & 16.71 & 8.03 & 25.26 & 20.17 & 24.88 & 60.00 & 83.85 & 33.14 & \textbf{10.17} & 92.08 & 50.00 & 42.03 & 32.64 \\

&~~~H2O & 25.42 & 8.70 & 20.98 & 20.25 & 19.26 & 9.67 & 24.11 & 22.97 & 24.24 & 44.00 & 92.10 & 41.24 & 7.93 & 96.48 & 59.18 & 48.33 & 35.30 \\

&~~~SnapKV & 28.53 & 11.33 & 25.00 & \textbf{25.86} & 21.13 & 13.05 & 25.05 & 23.13 & 24.44 & 66.00 & 92.32 & 41.56 & 8.70 & 96.23 & 60.24 & \textbf{50.31} & \underline{38.31} \\

&~~~PyraKV & 27.46 & 11.31 & 25.19 & 20.62 & 18.88 & 10.95 & 24.37 & 23.08 & 24.42 & 67.00 & 91.82 & \textbf{42.50} & 8.69 & \textbf{96.83} & 59.56 & 48.60 & 37.58 \\

&~~~\textbf{Ours} & \textbf{32.48} & \textbf{12.98} & \textbf{26.40} & 24.63 & \textbf{22.49} & \textbf{14.96} & \textbf{27.62} & \textbf{23.91} & \textbf{25.38} & \textbf{68.50} & \textbf{92.44} & 42.30 & 7.10 & 95.20 & \textbf{60.65} & 49.66 & \textbf{39.17} \\

\midrule

\multirow{19}{*}{\rotatebox[origin=c]{90}{\textbf{Mistral-7B-Instruct-v0.3}}}

&~~~Full KV & 29.20 & 41.58 & 53.11 & 49.27 & 39.01 & 28.58 & 31.95 & 25.67 & 23.89 & 76.00 & 88.59 & 47.50 & 5.50 & 98.00 & 51.97 & 46.28 & 46.01 \\
\cline{2-19}

& \multicolumn{18}{c}{\cellcolor{lightgray!25} 
  \textbf{KV Cache Size = 128}
} \\

&~~~StrLLM & 21.42 & 20.42 & 30.56 & 40.83 & 32.77 & 18.26 & 18.12 & 19.70 & 16.42 & 46.00 & 81.07 & 31.78 & 6.00 & 81.00 & 37.61 & 34.88 & 33.55 \\

&~~~H2O & 24.48 & 28.97 & 43.61 & 42.98 & 35.31 & 24.03 & 17.46 & 23.04 & 17.60 & 38.50 & 87.67 & 43.57 & \textbf{6.50} & \textbf{95.50} & 47.60 & 41.52 & 38.65 \\

&~~~SnapKV & 23.86 & 28.02 & 51.44 & 47.71 & 35.90 & 24.16 & 16.53 & 21.79 & 16.70 & 47.50 & 89.54 & 42.98 & 6.00 & 93.50 & 48.87 & \textbf{41.56} & \underline{39.75} \\

&~~~PyraKV & 24.21 & 27.04 & 49.20 & 47.52 & 34.81 & 23.51 & 16.78 & 21.39 & 16.87 & 48.00 & 88.16 & 43.05 & 5.50 & 94.00 & \textbf{49.08} & 40.93 & 39.38 \\

&~~~\textbf{Ours} & \textbf{27.86} & \textbf{35.71} & \textbf{54.33} & \textbf{48.14} & \textbf{38.69} & \textbf{26.97} & \textbf{20.43} & \textbf{24.23} & \textbf{19.96} & \textbf{56.50} & \textbf{90.16} & \textbf{46.29} & 5.50 & 93.00 & 48.74 & 41.55 & \textbf{42.38} \\

\cline{2-19}
& \multicolumn{18}{c}{\cellcolor{lightgray!25} 
  \textbf{KV Cache Size = 256}
} \\

&~~~StrLLM & 22.00 & 21.43 & 33.36 & 41.14 & 32.38 & 16.27 & 21.43 & 19.71 & 20.59 & 56.50 & 80.08 & 31.47 & \textbf{6.50} & 81.00 & 38.74 & 34.84 & 34.84 \\

&~~~H2O & 25.15 & 28.31 & 47.49 & 46.61 & 35.39 & 24.52 & 18.81 & \textbf{23.65} & 19.37 & 39.50 & 88.32 & 43.70 & 6.00 & \textbf{97.50} & 50.15 & 43.08 & 39.85 \\

&~~~SnapKV & 26.91 & 33.15 & 54.14 & 48.05 & 36.03 & 26.88 & 18.68 & 23.35 & 19.07 & 61.50 & 89.03 & 44.61 & 5.50 & 95.00 & \textbf{51.11} & 43.99 & \underline{42.31} \\

&~~~PyraKV & 25.67 & 31.84 & 54.15 & \textbf{48.43} & 36.38 & 26.62 & 18.76 & 23.35 & 19.07 & 62.00 & \textbf{89.46} & 44.42 & 6.00 & 96.00 & 50.88 & 44.00 & \underline{42.31} \\

&~~~\textbf{Ours} & \textbf{28.42} & \textbf{36.32} & \textbf{54.67} & 48.39 & \textbf{38.00} & \textbf{27.50} & \textbf{22.02} & 23.49 & \textbf{21.33} & \textbf{64.50} & 89.21 & \textbf{46.03} & \textbf{6.50} & 94.50 & 50.80 & \textbf{44.47} & \textbf{43.51} \\

\cline{2-19}
& \multicolumn{18}{c}{\cellcolor{lightgray!25} 
  \textbf{KV Cache Size = 512}
} \\

&~~~StrLLM & 22.22 & 22.98 & 37.29 & 40.29 & 32.89 & 16.47 & \textbf{24.58} & 19.98 & \textbf{23.00} & 65.50 & 80.33 & 31.06 & 5.50 & 83.50 & 40.35 & 35.90 & 36.37 \\

&~~~H2O & 25.61 & 31.24 & 50.52 & 46.62 & 36.14 & 24.19 & 19.91 & 24.36 & 21.02 & 41.50 & 89.18 & 45.67 & 5.50 & \textbf{97.50} & 51.56 & 44.43 & 40.93 \\

&~~~SnapKV & \textbf{28.46} & 35.53 & 55.18 & \textbf{49.30} & 37.40 & 27.59 & 21.03 & 23.87 & 20.75 & \textbf{70.50} & \textbf{89.57} & 44.72 & 5.50 & \textbf{97.50} & \textbf{52.01} & 45.67 & \underline{44.04} \\

&~~~PyraKV & 27.05 & 36.16 & \textbf{55.78} & 48.88 & 36.85 & 27.44 & 21.13 & 23.78 & 20.95 & \textbf{70.50} & 89.36 & 45.18 & 3.50 & 96.50 & 51.93 & 45.23 & 43.76 \\

&~~~\textbf{Ours} & 28.08 & \textbf{38.16} & 54.08 & 48.96 & \textbf{38.91} & \textbf{27.76} & 24.28 & \textbf{24.68} & 22.06 & 68.00 & 89.36 & \textbf{47.06} & \textbf{7.00} & 96.00 & 51.65 & \textbf{45.68} & \textbf{44.48} \\

\specialrule{1pt}{2pt}{0pt}
\end{tabular}

\end{threeparttable}

\normalsize\caption{Performance comparison of different methods on LongBench under different budgets across various LLMs. The best result is in bold and the second best result is underlined. The local window size of all baseline is 32, and the length of soft tokens in our method is also 32. In the method, StrLLM and PyraKV correspond to StreamingLLM and PyramidKV respectively.}
\label{tab:longbench}
\end{table*}

\subsection{Experimental Setting}

\paragraph{Training data and configuration.}
We train models using a subset extracted from ShareGPT dataset~\cite{ShareGPT-Chinese-English-90k}.
As described in Sec~\ref{subsec:training}, this subset comprises 50,000 samples, with 45,000 from the common domain and 5,000 from computer. 
Experimental validation indicates that choosing $ n = 32 $ provides an effective balance between training efficiency and generalization. 
We conduct experiments on Llama-3.1-8B-Instruct and Mistral-7B-Instruct-v0.3.

\paragraph{Baselines.}
To validate the effectiveness of our method, we select three commonly used and effective KV cache eviction methods as baselines.
During the evaluation, all baselines are tested under the same settings, i.e., local window size and pooling configurations, etc.
\begin{itemize}
    \item \textbf{StreamingLLM}~\cite{xiao2023efficient} heuristically retains the key-value pairs corresponding to the beginning of the input as well as those from the local window.
    \item \textbf{H2O}~\cite{zhang2023h2oheavyhitteroracleefficient} utilizes cumulative attention scores as importance scores. For ease of implementation, we adopt the approach outlined in~\cite{cai2024pyramidkv}, where the tokens from last window during the pre-filling stage are used as queries to compute importance scores.
    \item \textbf{SnapKV}~\cite{li2024snapkvllmknowslooking} calculates attention scores using tokens within the local window and applies a pooling mechanism to derive the final importance scores, ensuring consistency in KV cache eviction.
    \item \textbf{PyramidKV}~\cite{cai2024pyramidkv} builds on SnapKV by dynamically allocating budgets across layers, with the budget gradually decreasing as the layer depth increases.
\end{itemize}

\paragraph{Benchmarks.}

We evaluate methods using three common benchmarks for KV cache eviction: LongBench~\cite{bai2023longbench}, RULER~\cite{hsieh2024ruler}, and Needle-in-a-Haystack~\cite{kamradt2023needle}, and compare the performance of our method against baselines under multiple budgets.

\subsection{Main Results}
\label{subsec:results}
\paragraph{Results on LongBench.} 

As shown in Table~\ref{tab:longbench}, under the same budget, our method outperforms these KV cache eviction baselines by approximately 1 point and even by up to 2.6 points. 
Moreover, the lower the budget, the greater the improvement achieved by our method. 
In addition, in certain tasks, the performance of our method even surpasses that of Full KV. 
This demonstrates the effectiveness of our method.

\begin{table}
\setlength{\tabcolsep}{1mm}
\small
  \begin{tabular}{*{8}{c}}
  \toprule
  & \multirow{2}{*}[-0.5ex]{\textbf{Budget}} & \multicolumn{6}{c}{\textbf{Method}} \\
  \cmidrule(){3-8}
    & & Full KV & StrLLM & H2O & SnapKV & PyraKV & \textbf{Ours} \\
  \midrule
  \multirow{8}{*}{\rotatebox[origin=c]{90}{\textbf{Llama-3.1-8B-Ins}}}
  & \multicolumn{7}{c}{\cellcolor{lightgray!25} 
  \textbf{Sequence Length = 8192}
} \\
  & 256 & 87.18 & 14.45 & 28.71 & \underline{57.83} & 56.86 & \textbf{63.13} \\
  & 512 & 87.18 & 16.95 & 41.61 & \underline{62.76} & 61.19 & \textbf{69.24} \\
  & 1024 & 87.18 & 21.89 & 53.22 & \underline{68.21} & 66.30 & \textbf{74.12} \\
  \cline{2-8}
  & \multicolumn{7}{c}{\cellcolor{lightgray!25} 
  \textbf{Sequence Length = 32768}
} \\
  & 256 & 78.82 & 13.08 & 23.83 & \underline{53.51} & 53.25 & \textbf{56.31} \\
  & 512 & 78.82 & 14.00 & 34.02 & \underline{57.87} & 57.01 & \textbf{63.08} \\
  & 1024 & 78.82 & 16.29 & 43.56 & \underline{61.64} & 60.60 & \textbf{67.55} \\
  \midrule
  \multirow{8}{*}{\rotatebox[origin=c]{90}{\textbf{Mistral-7B-Ins-v0.3}}}
  & \multicolumn{7}{c}{\cellcolor{lightgray!25} 
  \textbf{Sequence Length = 8192}
} \\
  & 256 & 81.87 & 10.55 & 15.06 & \underline{42.44} & 40.35 & \textbf{52.72} \\
  & 512 & 81.87 & 12.81 & 21.28 & \underline{53.12} & 50.14 & \textbf{61.95} \\
  & 1024 & 81.87 & 17.98 & 30.09 & \underline{61.74} & 58.69 & \textbf{71.24} \\
  \cline{2-8}
  & \multicolumn{7}{c}{\cellcolor{lightgray!25} 
  \textbf{Sequence Length = 32768}
} \\
  & 256 & 73.46 & 9.95 & 11.32 & \underline{26.25} & 24.67 & \textbf{33.48} \\
  & 512 & 73.46 & 10.69 & 13.16 & \underline{33.05} & 31.69 & \textbf{40.83} \\
  & 1024 & 73.46 & 11.92 & 16.20 & \underline{40.63} & 39.13 & \textbf{49.97} \\
  \bottomrule
  \end{tabular}
    \normalsize\caption{Performance on RULER under different budgets across various LLMs. The number of samples in the test dataset is 100, and all other settings are the same as Table~\ref{tab:longbench}.}
  \label{tab:ruler}
\end{table}

\paragraph{Results on RULER.}

We evaluate the performance of the methods under different budgets with input lengths of 8192 and 32768, respectively. 
The experimental results are presented in Table~\ref{tab:ruler}. 
Our method significantly outperforms all baselines, consistently maintaining an advantage of over 3 points and and peaking at nearly 10 points. 
Additional results under other settings are provided in the repository.

\paragraph{Results on Needle-in-a-Haystack.}

\begin{figure}[!htbp]
    \centering
    \includegraphics[width=\linewidth]{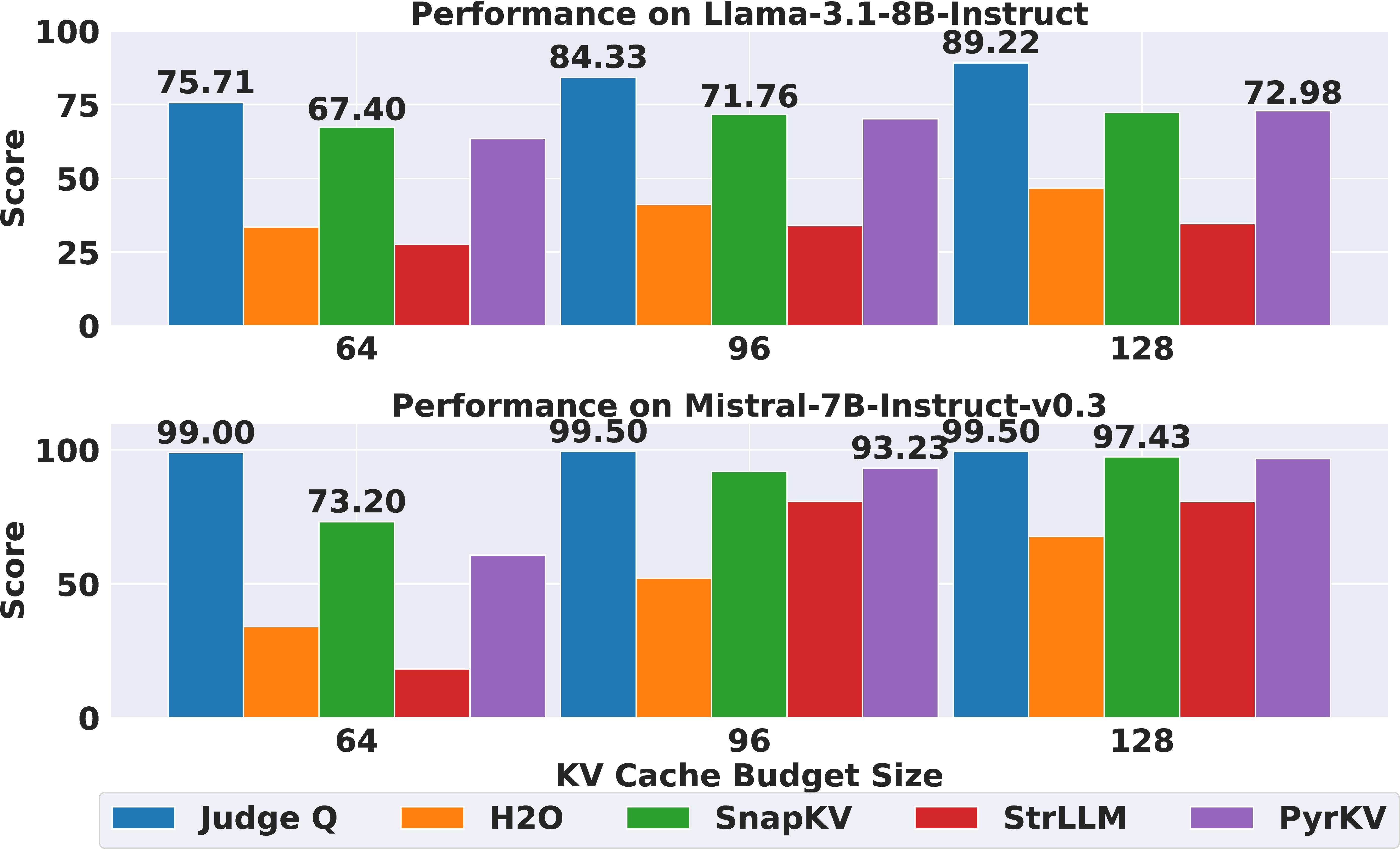}
    \caption{Performance on Needle-in-a-Haystack. The settings are the same as previous experiments. We set the minimum length to 4000, the maximum length to 32000, and the intervals for the document depth percent to 10.}
    \label{fig:nih}
\end{figure}

The experimental results are shown in Figure~\ref{fig:nih}. 
It is evident that our method significantly exceeds all baselines, indicating its robust performance in retrieval scenarios.
Detailed visualization results are available in the repository.

\section{Discussion}

\subsection{The Effect of Soft Tokens}
Based on the test results on benchmarks in Sec~\ref{subsec:results}, we can observe the effectiveness of our method. 
We further conduct experiments to explain the advantages of the soft tokens.
\paragraph{Soft tokens can improve the critical key-value hit rate.} 
As discovered in Section~\ref{sec:obs}, utilizing actual decoded tokens to guide KV cache eviction can yield excellent results, and it can be considered the theoretical upper bound of KV cache eviction methods. 
To evaluate this, we introduce a new metric called the critical key-value hit rate. 
As defined in Equation~\ref{eq:hit_rate}, this metric measures the overlap between the key-value pairs selected by the current eviction method and those chosen using real decoded tokens under the same budget. 
By measuring the critical key-value hit rate, we can directly compare the discrepancy between the current eviction method and the theoretical upper bound.

\begin{equation}
\label{eq:hit_rate}
\operatorname{HitRate}(\mathbf{I}_{\text{cur}}, \mathbf{I}_{\text{resp}})
=
\frac{\left|\, \mathbf{I}_{\text{cur}} \cap \mathbf{I}_{\text{resp}} \,\right|}
     {\left|\, \mathbf{I}_{\text{resp}} \,\right|}
\end{equation}

According to the experimental results in Sec~\ref{subsec:results}, SnapKV performs the best among all baselines. 
Therefore, we only compare the hit rate of Judge Q and SnapKV under different budgets. 
We preform evaluation using Llama-3.1-8B-Instruct on LongBench.
As shown in Table~\ref{tab:hit_rate}, the critical key-value hit rate of our method consistently exceeds that of SnapKV by approximately 8 points across various budgets. 

This indicates that during the KV cache eviction process, our method is more effective at retaining important key-value pairs that meet decoding requirements, preserving more information and enhancing model's performance.
It is also consistent with the training objective of Judge Q, which is to train soft tokens to identify key-value pairs that closely approximate those selected by actual decoded tokens.
\begin{table}[!htbp]
\centering
  \begin{tabularx}{\linewidth}{*{4}{>{\centering\arraybackslash}X}}
  \toprule
  \multirow{2}{*}[-0.5ex]{\textbf{Method}} & \multicolumn{3}{c}{\textbf{KV Cache Budget}} \\
  
  \cmidrule(){2-4}
  & 128 & 256 & 512 \\
  \midrule
   SnapKV & 53.44 & 55.23 & 58.46 \\
  \textbf{Ours} & \textbf{61.37} & \textbf{62.34} & \textbf{65.06} \\
  \bottomrule
  \end{tabularx}
    \caption{Critical key-value hit rate of our method and SnapKV on LongBench under different budgets.}
  \label{tab:hit_rate}
\end{table}

\begin{figure*}[t]
    \centering
    \begin{subfigure}{0.48\linewidth}
        \includegraphics[width=\linewidth]{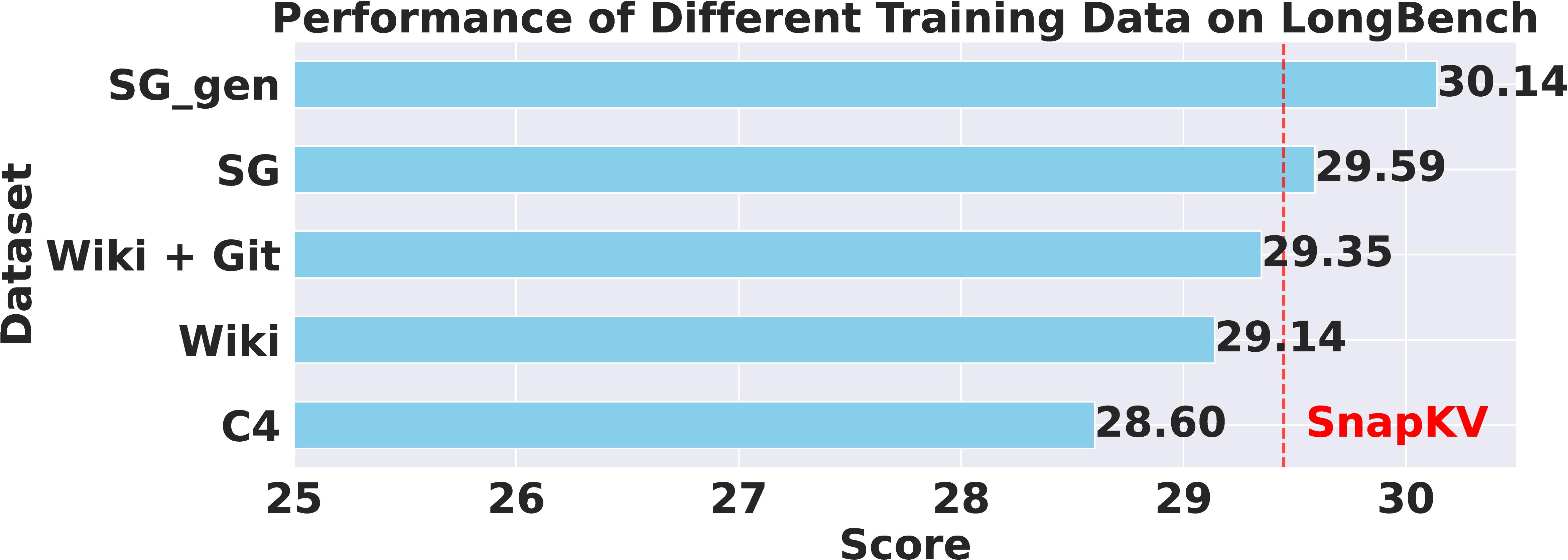}
        \caption{Performance of Llama-3.2-1B-Instruct on LongBench across training datasets (SG = ShareGPT, Wiki = Wikipedia\_en, Git = GitHub repos), using consistent setup and a budget of 512.}
        \label{fig:dataset}
    \end{subfigure}
    \hfill
    \begin{subfigure}{0.48\linewidth}
        \includegraphics[width=\linewidth]{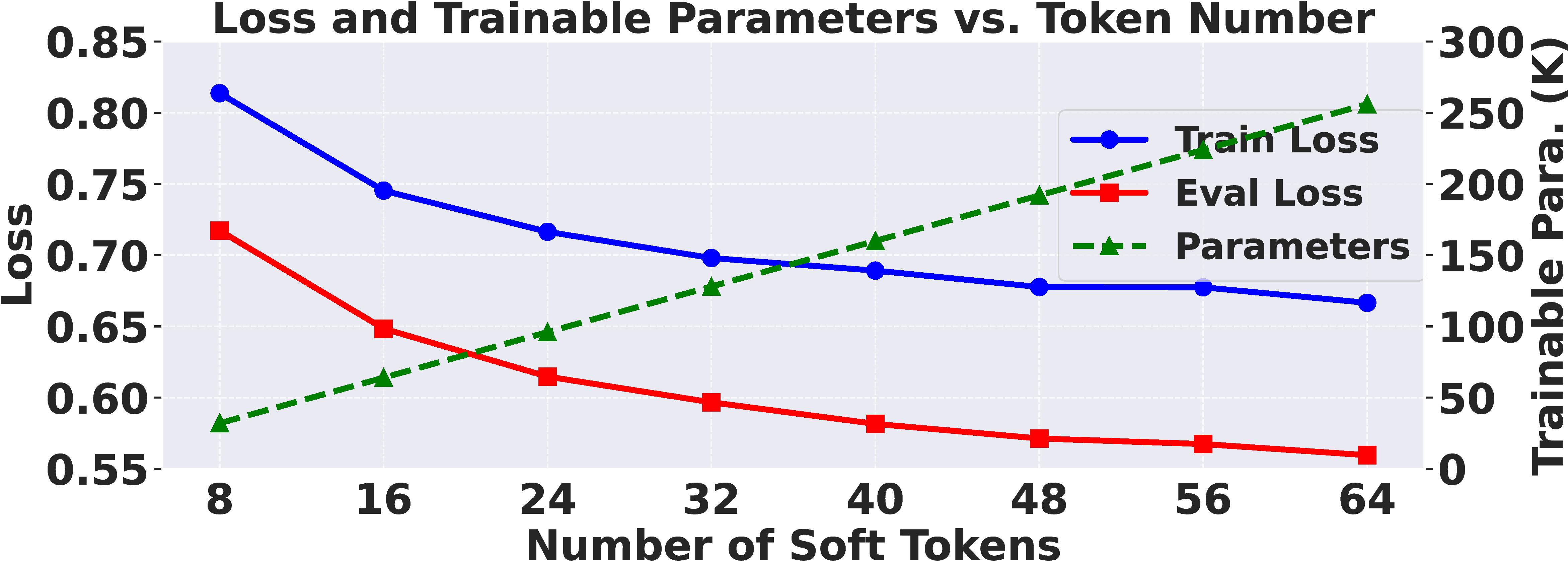}
        \caption{Loss and trainable parameters under different numbers of soft tokens during training on Llama-3.1-8B-Instruct. The hyperparameters used for training are the same as prior experiments.}
        \label{fig:soft_tokens}
    \end{subfigure}
    \caption{The impact of data quality and the number of soft tokens.}
    \label{fig:combined}
\end{figure*}

\paragraph{Using soft tokens for eviction can ensure the quality of generated content.} 
We further evaluate the effectiveness of Judge Q. 
Following the prompt template adjustment settings described in Section~\ref{sec:obs}, we test the proportion of performance degradation when the question is placed at the start of the prompt. 
Under the same settings, our method exhibits less than 7\% performance loss, which is lower compared to the 
approximately 10\% loss observed in baselines. 
\begin{table}[!htbp]
\centering
  \begin{tabularx}{0.8\linewidth}{*{3}{>{\centering\arraybackslash}X}}
  \toprule
  \multirow{2}{*}[-0.5ex]{\textbf{Method}} & \multicolumn{2}{c}{\textbf{Dataset}} \\
  \cmidrule(){2-3}
     & MATH-500 & AIME24\\
  \midrule
   SnapKV & 52.4 & 31.1 \\
  \textbf{Ours} & \textbf{55.0} & \textbf{37.8} \\
  \bottomrule
  \end{tabularx}
    \caption{Performance on text continuation tasks, under a budget of 25\% of average output length. Specifically, the budget is set to 1024 for MATH-500 and 3072 for AIME24.}
  \label{tab:reasoning}
\end{table}

Additionally, we conduct experiments on text continuation tasks. 
Specifically, we apply the Judge Q framework to DeepSeek-R1-Distill-Llama-8B~\cite{deepseekai2025deepseekr1incentivizingreasoningcapability} and evaluate its performance on the MATH-500~\cite{lightman2023lets} and AIME24~\cite{AIME2024} datasets. 
For each dataset, the model initially generates responses for individual questions. 
Subsequently, we extract the last 768 tokens from the generated responses, and perform KV cache eviction before continuing to generate responses. 
Finally, we evaluate the results of the regenerated outputs.
As shown in Table~\ref{tab:reasoning}, our method exhibits strong performance in KV cache eviction under long-output scenarios. 
Notably, in the case of AIME24, where the output exceeds 10,000 tokens and only one-quarter of the tokens are retained, our method significantly outperforms SnapKV.

\newtcolorbox{examplebox}{
    colback=blue!3!white,  
    colframe=blue!70!black, 
    fonttitle=\bfseries,
    title=Case Study,
    boxrule=1pt,
    top=3pt,
    left=6pt,
    right=6pt,
    bottom=3pt,
    before skip=5pt,     
    after skip=5pt,       
    boxsep=3pt             
}

\newcommand{\question}[1]{\textcolor{blue}{\textbf{Prompt:}} #1\par\vspace{-3pt}}
\newcommand{\snapresponse}[1]{\textcolor{orange!70!black}{\textbf{SnapKV Response:}} #1 \textcolor{red}{$\times$} \par\vspace{-3pt}}
\newcommand{\judgeresponse}[1]{\textcolor{teal!70!black}{\textbf{Judge Q Response:}} #1 \textcolor{green!60!black}{$\checkmark$} \par\vspace{-3pt}}
\newcommand{\predresponse}[1]{\textcolor{cyan!60!black}{\textbf{Full KV Response:}} #1 \textcolor{green!60!black}{$\checkmark$} \par\vspace{-3pt}}
\newcommand{\goldresponse}[1]{\textcolor{green!60!black}{\textbf{Gold Response:}} #1}

\begin{examplebox}
    \question{You are given a story, which can be either a novel or a movie script, ... Question: What is Saltram's living situation? Answer: (31406 tokens)}\par\medskip
    
    \snapresponse{He is living with the Saltrams, who have a large house.}\par\medskip
    
    \judgeresponse{He is living with the Mulvilles in their drawing room.}\par\medskip

    \predresponse{He is living with the Mulvilles at Wimbledon.}\par\medskip
    
    \goldresponse{He is a guest in the home of the Mulvilles.}
\end{examplebox}

Furthermore, we analyze specific case to compare the differences in the generated content between Judge Q and other eviction methods.
We select an example from Narrative QA using Llama-3.1-8B-Instruct as follows with a KV cache budget set to 512. 
It is evident that the response generated by Judge Q closely resembles the of Full KV and correctly answers the question, whereas SnapKV deviates significantly from the Full KV output, resulting in an incorrect response.
This validates that Judge Q can effectively maintain high generation quality.

\subsection{The Impact of Data Quality}

\paragraph{Training with model-generated responses yields better results.} 
The quality of training data can impact the effectiveness of our method to some extent.
During experiments, we observe that using model-generated responses outperforms directly using the original responses from the datasets. 
For efficiency, we conduct data quality explorations using Llama-3.2-1B-Instruct and evaluate on LongBench.
From results in Figure~\ref{fig:dataset}, it is evident that utilizing model-generated responses as training data is more effective.

\paragraph{Training data content crucial for performance.} 

We also extract and combine training data from various pre-training datasets, including C4~\cite{2020t5}, Wikipedia~\cite{wikidump} and GitHub. 
As Figure~\ref{fig:dataset} shows, these results are inferior to ShareGPT, because the data content is limited and does not cover a comprehensive range of domains.

\subsection{The Impact of the Number of Soft Tokens}

Additionally, during training, we construct a validation set comprising 5000 samples to explore the impact of the number of soft tokens. 
As illustrated in Figure~\ref{fig:soft_tokens}, when the number of soft tokens is around 32, there is an optimal balance between training cost and generalization performance.

\section{Conclusion}

In this paper, we first identify that existing KV cache eviction methods during the pre-filling stage, which directly select the local window for importance calculation, tend to overly focus on local information while neglecting global information. 
To address this issue, we propose Judge Q, a lightweight training framework that introduces a small number of soft tokens into the vocabulary, and only fine-tunes a small portion of the model's embedding parameters.
This approach enhances the consideration of global information. 

We conduct experiments using various models across multiple benchmarks, demonstrate that our method consistently outperforms existing KV cache eviction methods under different budgets.
Judge Q outperforms existing methods by approximately 1 point on LongBench, exceeds by more than 3 points on RULER, and significantly surpasses the baselines on Needle-in-a-Haystack. 
We further validate the effect of soft tokens as well as the impact of data quality and the number of soft tokens.

In the future, we plan to conduct experiments with a wider range of models, use longer data for training and implement streaming KV cache eviction during the decoding process.

\section * {Acknowledgments}
We gratefully acknowledge the support of the National Natural Science Foundation of China (NSFC) via grant 62236004, 62206078 and 62476073.

\bibliography{all}

\end{document}